\documentclass[10pt,twocolumn,letterpaper]{article}

\usepackage{iccv}
\usepackage{times}
\usepackage{epsfig}
\usepackage{graphicx}
\usepackage{amsmath}
\usepackage{amssymb}
\usepackage{multirow}
\usepackage{array}
\usepackage{upgreek}
\usepackage{color}
\usepackage{siunitx}
\usepackage[T1]{fontenc}
\usepackage{url}
\usepackage[pagebackref=true,breaklinks=true,letterpaper=true,colorlinks,bookmarks=false]{hyperref}

\iccvfinalcopy %

\usepackage[margin=4pt,font=footnotesize,labelfont=bf,labelsep=endash,tableposition=top]{caption}

\ificcvfinal\fi
\begin{document}

\title{Exploiting temporal consistency for real-time video depth estimation\thanks{Work was done when H. Zhang was visiting University of Adelaide. Correspondence should be addressed to C. Shen and Y. Li.
}
}

\author{
Haokui Zhang$ ^1$,
~~
Chunhua Shen$ ^2$,
~~
Ying Li$ ^1$,
~~
Yuanzhouhan Cao$ ^2$,
~~
Yu Liu$ ^2$, 
~~
Youliang Yan$ ^3$\\
$ ^1$Northwestern Polytechnical University~~
$ ^2$University of Adelaide~~
$ ^3$Noah's Ark Lab, Huawei 
}

\maketitle
\thispagestyle{empty}

\begin{abstract}
Accuracy 
of depth estimation from static images has been significantly improved recently,
by exploiting hierarchical features from deep convolutional neural networks (CNNs). Compared with static images, vast information exists among video frames and can be exploited to improve the depth estimation performance. In this work, we focus on exploring temporal information from monocular videos for depth estimation. Specifically, we take the advantage of convolutional long short-term memory (CLSTM) and propose a novel spatial-temporal CSLTM (ST-CLSTM) structure. Our ST-CLSTM structure can capture not only the spatial features but also the temporal correlations/consistency  among consecutive video frames with negligible increase in computational cost.  Additionally, in order to maintain the temporal consistency among the estimated depth frames, we apply the generative adversarial learning scheme and design a temporal consistency loss. The temporal consistency loss is combined with the spatial loss to update the model in an end-to-end fashion. 
By taking advantage of the
temporal information, we %
build a video depth estimation framework that runs in real-time and generates visually pleasant results. Moreover, our approach is %
flexible and can be generalized to most existing depth estimation frameworks. Code is available at:
\url{https://tinyurl.com/STCLSTM}

\end{abstract}

\tableofcontents

\section{Introduction}

Benefiting from the powerful convolutional neural networks (CNNs), some recent methods~\cite{zhou2018unsupervised,fu2018deep,mahjourian2018unsupervised,eigen2014depth,laina2016deeper} have achieved outstanding performance on depth estimation from monocular static images. The success of these methods is based on the deeply stacked network structures and large amount of training data. For instance, the state-of-the-art depth estimation model DORN~\cite{fu2018deep} has more than one hundred of convolution layers, the high computational cost may hamper it from practical applications. However, in some scenarios such as automatic driving~\cite{borghi2017poseidon} and robots navigation~\cite{mirowski2016learning}, estimating of depths in real-time is required. Directly extend existing methods from static image to video sequence is not feasible because of the 
excessive 
computational cost. In addition, sequential frames which contain rich temporal information are usually provided in such scenarios. The existing methods fail to take the temporal information into consideration.

\begin{figure*}
\begin{center}
\includegraphics[width=6.8in]{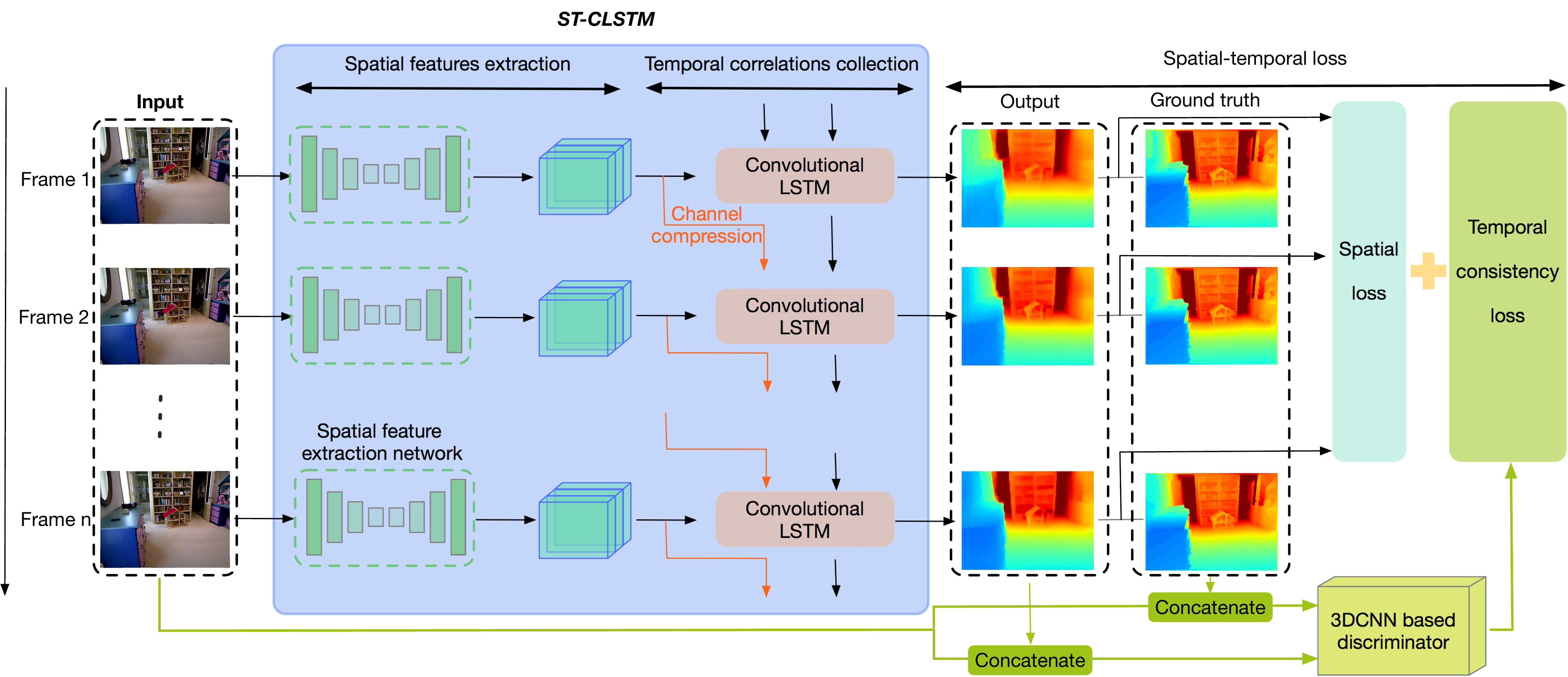}
\end{center}
   \caption{Illustration of our framework. The framework contains three main parts: spatial features extraction; temporal correlations collection; and spatial-temporal loss calculation. The first two parts 
   consist of 
   our ST-CLSTM structure which captures both spatial features and temporal correlations. After the ST-CLSM generates depth estimations, a 3D CNN is introduced to calculate the temporal loss. The spatial and temporal losses are combined to update the framework.}

\label{fig:framework}

\end{figure*}

In this work, we exploit temporal information from videos by making use of the convolutional long short-term memory (CLSTM) and the generative adversarial networks (GANs), and propose a real-time depth estimation framework. We illustrate our proposed framework in Fig.~\ref{fig:framework}. It consists of three main parts: 1) spatial features extraction part; 2) temporal correlations collection part and 3) spatial-temporal loss calculation part.
The spatial features extraction part and the temporal correlations collection part compose our novel spatial-temporal CLSTM (ST-CLSTM) structure. The spatial features extraction part first takes as input $n$ continuous frames $\left({x}^{1}, {x}^{2}, \cdots ,{x}^{n}\right)$ and outputs high level features $\left({f}^{1}, {f}^{2}, \cdots ,{f}^{n}\right)$. The temporal correlations collection part  then takes as input the high-level features and outputs depth estimations $\left({d}^{1}, {d}^{2}, \cdots ,{d}^{n}\right)$. With the \textit{cell} and \textit{gate} modules, the CLSTM can make use of the cues acquired from the previous frame to reason the current frame, and thus encode the temporal information. As for spatial-temporal loss calculation, we first calculate the spatial loss between the estimated and the ground-truth depths. In order to further enforce the temporal consistency, we design a new temporal loss by introducing a generative adversarial learning scheme. Specifically, we apply a 3D CNN as the discriminator which takes as input the estimated and ground-truth depth sequences and outputs the temporal loss. The temporal loss is combined with the spatial loss and back propagated through the entire framework to update the weights in an end-to-end fashion.

To %
summarize, 
our main contributions  are as follows.
\begin{itemize}
\itemsep -4pt
\item We propose a novel ST-CLSTM structure that is able to capture spatial features as well as temporal correlations for video depth estimation. To our knowledge, this is the first time that %
CLSTM is employed for video depth estimation.

\item We design a novel temporal consistency loss by %
using 
the generative adversarial learning scheme. Our temporal loss can further enforce the temporal consistency and improve the performance for video depth estimation.

\item Our proposed video depth estimation framework can execute  in real-time and can be generalized to most existing depth estimation frameworks.

\end{itemize}

\subsection{Related work}

{\bf Depth estimation}
Recently, 
many  deep learning based depth estimation methods have been proposed and achieved significant achievements. To name a few, Eigen \emph{et al.}~\cite{eigen2014depth} employed a multi-scale neural network with two components to generate coarse estimations globally and refine the results locally. Xie \emph {et al.}~\cite{xie2016deep3d} used shortcut connections in their network to fuse low-level and high-level features. Cao \emph{et al.}~\cite{cao2016estimating} proposed to formulate depth estimation as a classification problem instead of a regression problem. Laina \emph{et al.}~\cite{laina2016deeper} employed a reverse huber loss to estimate depth distributions and an up-sampling module to overcome the low-resolution problem. Yin \emph{et al.}~\cite{Yin2019enforcing} designed a loss term to enforce geometric constraints. To further improve the performance, some methods incorporate conditional random fields in their methods~\cite{wang2015towards, liu2016learning}. 
Recently the method DORN~\cite{fu2018deep} proposed a spacing-increasing discretization (SID) policy and estimated depths with a ordinal regression loss. Although excellent performance has been achieved, the networks are  deep and 
 computation is heavy.

Some other works focus on estimating depth values from videos. Zhou \emph{et al.}~\cite{zhou2018unsupervised} proposed to use  bundle adjustment as well as a super-resolution network to improve  depth estimation. Specifically, the bundle adjustment is used to estimate depths and camera poses simultaneously, and the super-resolution network is used to recover details. Mahjourian \emph{et al.}~\cite{mahjourian2018unsupervised} incorporated a 3D loss with geometric constraints to estimate depths and ego-motions simultaneously.
In this work, we propose to estimate depths by exploiting temporal information from videos.

\textbf{CLSTM in video analysis}
Recurrent neural networks (RNNs), especially the long short-term memories (LSTMs) have achieved great success in various computer vision tasks such as language processing~\cite{mikolov2011extensions} and speech recognition~\cite{graves2013speech}. With the memory cells, LSTMs can capture short and long term temporal dependencies. However, conventional LSTMs only take as input one-dimensional vectors and thus can not be applied to image sequence processing.

To overcome this limitation, Shi \emph{et al.}~\cite{xingjian2015convolutional} proposed convolutional LSTM (CLSTM), which can capture long and short term temporal dependencies while retaining the ability of handling two-dimensional feature maps. Recently, CLSTMs have been used in video processing. In~\cite{song2018pyramid}, Song \emph{et al.} proposed a Deeper Bidirectional CLSTM (DB-CLSTM) structure which learns temporal characteristics in a cascaded and deeper way for video salient object detection. Liu \textit{et al.}~\cite{liu2016spatio} proposed a tree-structure based traversal method to model the 3D-skeleton of a human being in spatial-temporal domain. They applied CLSTM to handle the noise and occlusions in 3D skeleton data, which improves the temporal consistency of the results. Jiang \emph{et al.}~\cite{jiang2017predicting} developed a two-layer ConvLSTM (2C-LSTM) to predict video saliency. An object-to-motion convolutional neural network has also been proposed.

\textbf{GAN}
The generative adversarial network (GAN) has been an active
research topic since it was proposed by Goodfellow \emph{et al.}~in \cite{goodfellow2014generative}. The basic idea of GAN is the training of two adversarial networks, a generator and a discriminator. During the process of adversarial training, both generator and discriminator become more robust. GANs have been widely used in various applications, such as image-to-image translation~\cite{isola2017image} and synthetic data generation~\cite{mahmood2018unsupervised}.
GAN has been mainly used for generating images. One of the first work to apply adversarial training to improve structured output learning might be \cite{PoseNet},
where a discriminator loss is used to distinguish predicted pose and ground-truth pose for pose estimation from monocular images. 
Recently, GANs have also been adopted in depth estimation. In~\cite{almalioglu2018ganvo}, Almalioglu \emph{et al.}\ employed GAN to generate sharper and more accurate depth maps. 

In this paper, we design a novel temporal loss by 
employing 
GAN. Our temporal loss can enforce the temporal consistency among video frames. 

\section{Our Method}

In this section, we elaborate on our proposed video depth estimation framework. We first introduce our ST-CLSTM structure; then we present our generative adversarial learning scheme and our spatial and temporal loss functions.

\subsection{ST-CLSTM}
Our depth estimation framework contains three main components: spatial feature extraction;
temporal correlation collection;
and  spatial-temporal loss calculation, as illustrated in Fig.~\ref{fig:framework}. 

\subsubsection{Spatial feature extraction network}

\begin{figure}
\begin{center}
\includegraphics[width=3.2in]{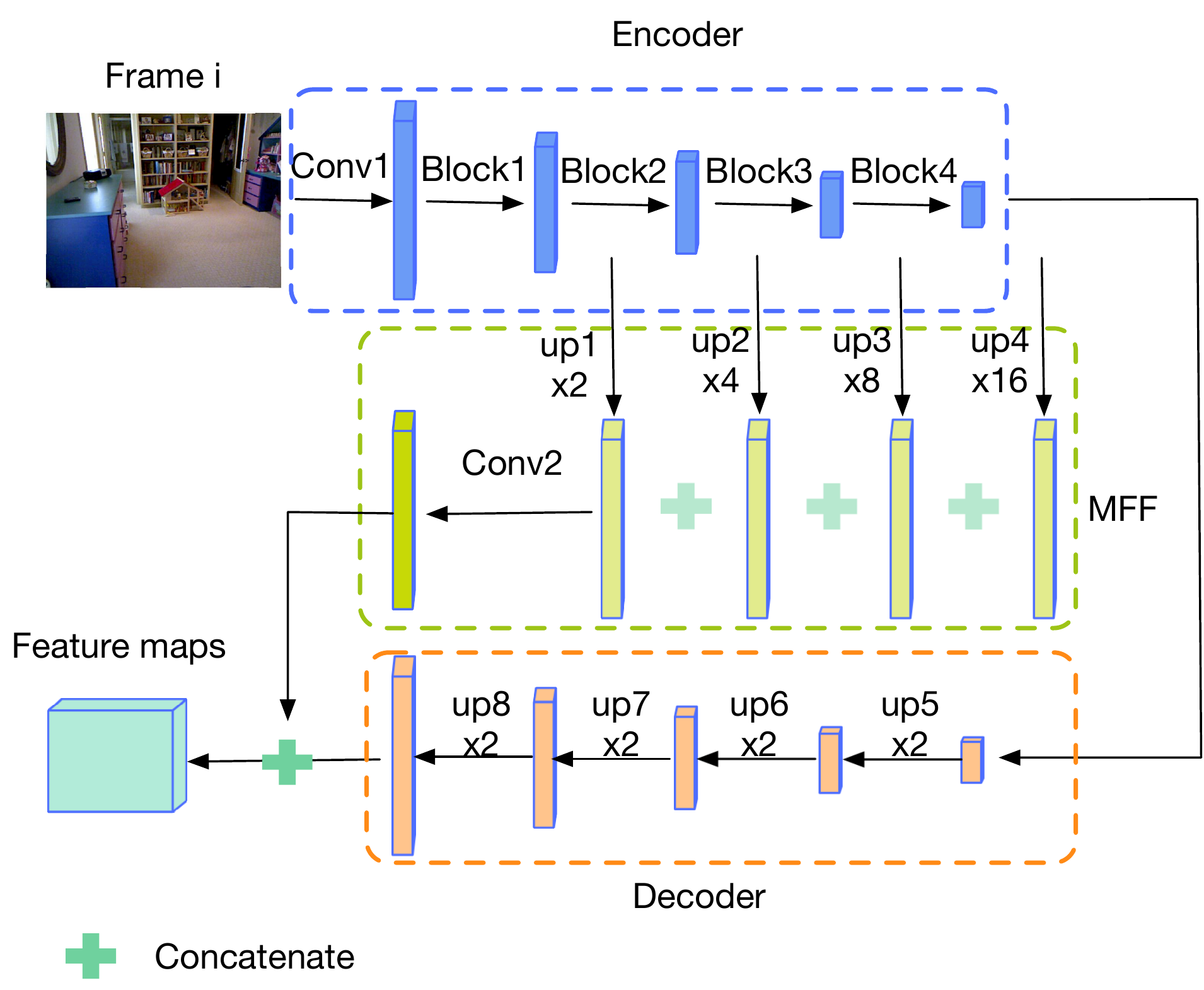}
\end{center}
   \vspace{-0.2cm}
   \caption{Spatial feature extraction network. This network consists of three parts, including an encoder, a decoder and a multi-scale feature fusion module (MFF). In this paper, we employ the relatively shallow model ResNet-18 as the encoder for fast processing.}
\label{fig:2DCNN}
\end{figure}

Spatial feature extraction %
is the key to the performance and processing speed as it contains the majority of trainable parameters in our depth estimation framework. In our work, we  use  a modified structure proposed by Hu~\emph{et al.}~\cite{hu2019revisiting}.

We show the details of our spatial feature extraction network 
in Fig.~\ref{fig:2DCNN}. The network contains  an encoder, a decoder and a multi-scale feature fusion module (MFF). The encoder can be any 2D CNN model, such as the VGG-16~\cite{simonyan2014very}, the ResNet~\cite{he2016deep}, the SENet~\cite{hu2018squeeze},  among many others. 
In order to build a real-time depth estimation framework, we apply a shallow ResNet-18 model instead of the SENet-154 as the encoder.

The decoder employs four up-projection modules to improve the spatial resolution and decreases the number of channels of the feature maps. This   encoder-decoder structure has been widely used in pixel-level tasks~\cite{chen2018encoder, fu2018deep}. The MFF module is designed to integrate features of different scales. Similar strategies are used  in~\cite{long2015fully}.

Note that, in our depth estimation framework, the spatial feature extraction network can be replaced by other depth estimation models. In other words, our proposed depth estimation framework can be applied to other state-of-the-art depth estimation methods with minimum modification.

\subsubsection{CLSTM}

\begin{figure}
\begin{center}
\includegraphics[width=3.0in]{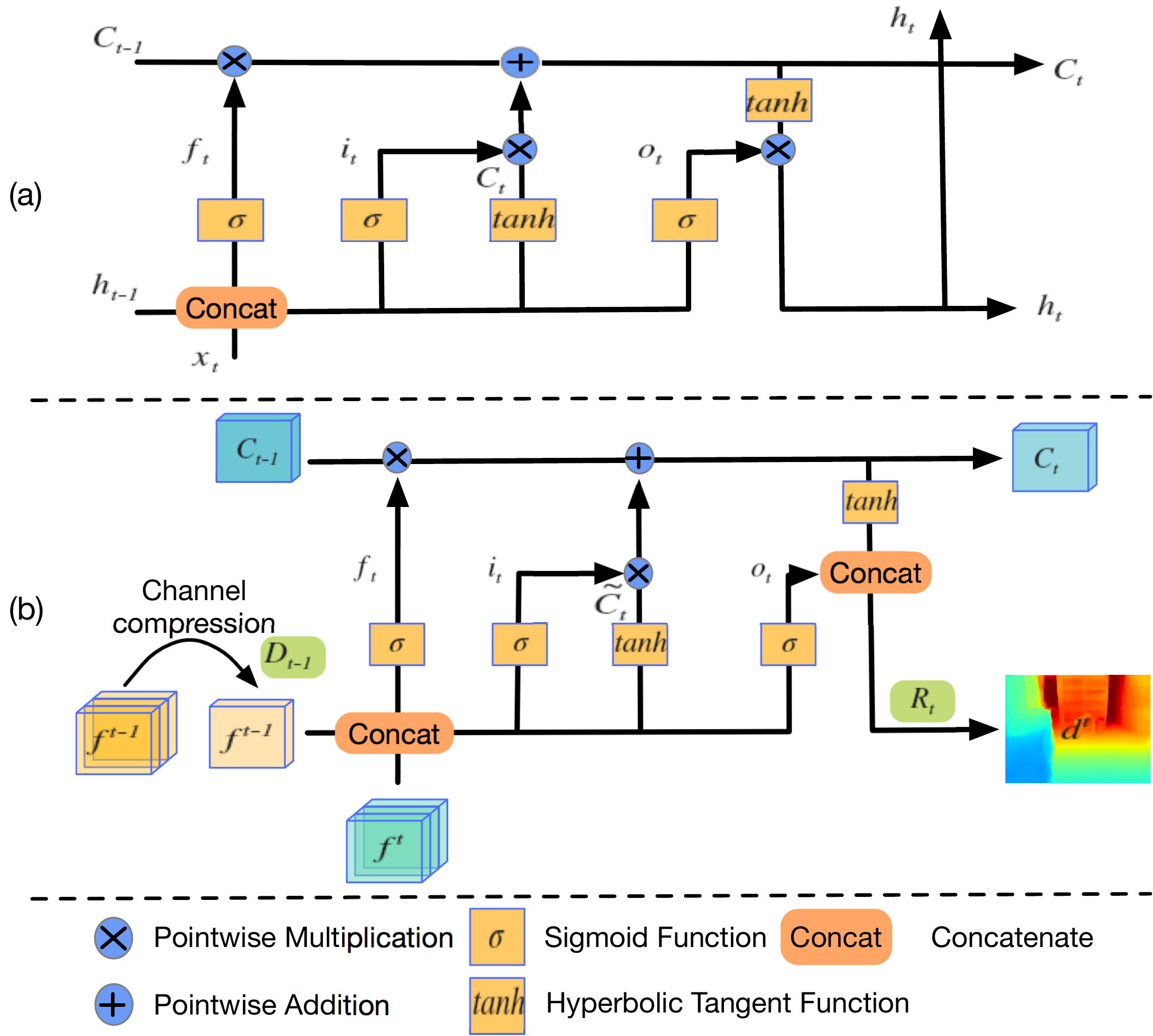}
\end{center}
   \caption{LSTM and CLSTM. (a) LSTM; (b) CLSTM. In LSTM, both the inputs and the outputs are vectors. In our proposed CLSTM, the inputs are feature maps and the the outputs are the estimated depths.}
\label{fig:Conv_LSTM}
\end{figure}

As the input frames are continuous in the temporal dimension, taking  the temporal correlations of these frames 
into consideration 
is intuitive and presumably helpful 
for improving depth estimation performance.  
In terms of achieving this goal, both the 3D CNN and the CLSTM are competent. Here, we use the CLSTM, as the it is more flexible than the 3D CNN for online inference. The structure of our proposed CLSTM is shown in Fig.~\ref{fig:Conv_LSTM} (b).

Fig.~\ref{fig:Conv_LSTM} (a) shows the traditional LSTM. The inputs and the outputs are vectors and the key operation is the Hadamard product. A single LSTM cell at time $t$ can be expressed as:
\begin{equation}
\begin{aligned}
&{f}_{t}=\sigma \left({{W}_{f}\circ \left[{h}_{t-1}, {x}_{t} \right] + {b}_{f}} \right), \\ 
&{i}_{t}=\sigma \left({W}_{i}\circ \left[{h}_{t-1}, {x}_{t} + {b}_{i} \right]  \right),  \\ 
&\tilde {{C}_{t}} = tanh \left({W}_{C}\circ \left[{h}_{t-1}, {x}_{t} \right]  + {b}_{C} \right), \\
&{C}_{t} = {f}_{t}\times {C}_{t-1} + {i}_{t}\times {\tilde{C}}_{t}, \\
&{o}_{t} = \sigma \left( {W}_{o}\circ \left[{h}_{t-1}, {x}_{t} \right] + {b}_{o} \right), \\
&{h}_{t} = {o}_{t} \times tanh\left({C}_{t} \right),
\end{aligned}
\end{equation}
where $\sigma$ and $tanh$ are sigmoid and hyperbolic tangent activation functions. $\circ$ and $\times$ represent the Hadamard product and pointwise multiplication.

Compared with the traditional LSTM, our proposed CLSTM exhibits  two main differences: 1) Operation. Following~\cite{xingjian2015convolutional}, we replace the Hadamard product in LSTM with convolution to handle the extracted 2D feature maps. 2) Structure. We adjust the structure of CLSTM to deal with depth estimation task. Specifically, our proposed CLSTM cell can be expressed as:
\begin{equation}
\begin{aligned}
&{f}_{t}=\sigma \left( \left[{f}^{t}, {D}_{t-1}({f}^{t-1}) \right]\ast {W}_{f} + {b}_{f}\right), \\
&{i}_{t}=\sigma([{f}^{t}, {D}_{t-1}({f}^{t-1})]\ast {W}_{i} + {b}_{i}), \\
&\tilde{{C}_{t}}=tanh([{f}^{t}, {D}_{t-1}({f}^{t-1})]\ast {W}_{C} + {b}_{C}), \\
&{C}_{t}={f}_{t}\times {C}_{t-1} + {i}_{t}\times \tilde{{C}_{t}}, \\
&{o}_{t}=\sigma([{f}^{t}, {D}_{t-1}({f}^{t-1})]\ast {W}_{o} + {b}_{o}), \\ 
&{R}_{t} = Conv([{o}_{t}, tanh({C}_{t})]),
\end{aligned}
\end{equation}
where $\ast$ is the convolutional operator. ${{W}_{f}, {W}_{i}, {W}_{C}, {W}_{o}}$ and ${{b}_{f}, {b}_{i}, {b}_{C}, {b}_{o}}$ denote the kernels and bias terms at the corresponding convolution layers. After we extract the spatial features of video frames, we feed the feature map of the previous frame ${f}^{t-1}$ into a convolution layer ${D}_{t-1}$ to compress the number of channels from $c$ to 8. Then we concatenate ${f}^{t-1}$ with the feature map of current frame ${f}^{t}$ to formulate a feature map with $c+8$ channels. Next, we feed the concatenated feature map to CLSTM to update the information stored in memory cell. Finally, we concatenate the information in the updated memory cell ${C}_{t}$ and the feature map of output gate, then feed them to a refine structure ${R}_{t}$ that consists of two convolution layers to obtain the final estimation result.

\subsection{Spatial-temporal loss}
As shown in Fig.~\ref{fig:framework}, the output of our ST-CLSTM is the estimated depth. We design two loss functions to train our ST-CLSTM model: a spatial loss to maintain the spatial features and a temporal loss to capture the temporal consistency. 

\subsubsection{Spatial loss}
We follow~\cite{hu2019revisiting} and design a similar loss function as our spatial loss, which can be expressed as:
\begin{equation}
{L}_{spatial}={l}_{depth}+\lambda{l}_{grad}+\mu{l}_{normal},
\end{equation}
where $\lambda$ and $\mu$ are weighting coefficients. It is composed of three terms. The ${l}_{depth}$ is applied to penalize inaccurate depth estimations. Most existing depth estimation methods simply apply the $\ell_1$ or $\ell_2$ loss. As pointed in~\cite{JHLee2018}, a problem of this type of loss is that the value tends to be larger as the ground-truth depth getting further. We apply a logarithm $\ell_1$ loss which is expressed as:
\begin{equation}
F(x,y)=\ln({||x-y||}_{1}+1.0).
\end{equation}
Consequently, our ${l}_{depth}$ is defined as:
\begin{equation}
{l}_{depth}=\frac{1}{n}\sum_{i=1}^{n}{F({d}_{i}, {g}_{i})},
\end{equation}
where $n$ is the number of pixels;
${d}_{i}$ and ${g}_{i}$ are the estimated and ground-truth depth of pixel $i$ respectively.

${l}_{grad}$ is designed to penalize the errors around edges. It is defined as:
\begin{equation}
\begin{aligned}
& {l}_{grad}= \\
& \frac{1}{n}\sum_{i=1}^{n}({F({\triangledown}_{x}({d}_{i}), {\triangledown}_{x}({g}_{i}))+F({\triangledown}_{y}({d}_{i}), {\triangledown}_{y}({g}_{i}))}),
\end{aligned}
\end{equation}
where ${\triangledown}_{x}$ and ${\triangledown}_{y}$ represent the spatial derivative along the  $x$-axis and $y$-axis respectively. 

The last item ${l}_{normal}$ is designed to measure the angle between two surface normals, and thus is sensitive to small depth structures. It is expressed as:
\begin{equation}
{l}_{normal}=\frac{1}{n}\sum_{i=1}^{n}{\left(1-\frac{{\eta}_{i}^{d}\cdot{\eta}_{i}^{g}}{\sqrt{{\eta}_{i}^{d}\cdot{\eta}_{i}^{d}} \sqrt{{\eta}_{i}^{g}\cdot{\eta}_{i}^{g}}}\right)},
\end{equation}
where ${\eta}_{i}^{d}=\left[-{\triangledown}_{x}({d}_{i}), -{\triangledown}_{y}({d}_{i}), 1 \right]$ and $\cdot$ denotes inner product.

\begin{figure}
\begin{center}
\includegraphics[width=1.8in]{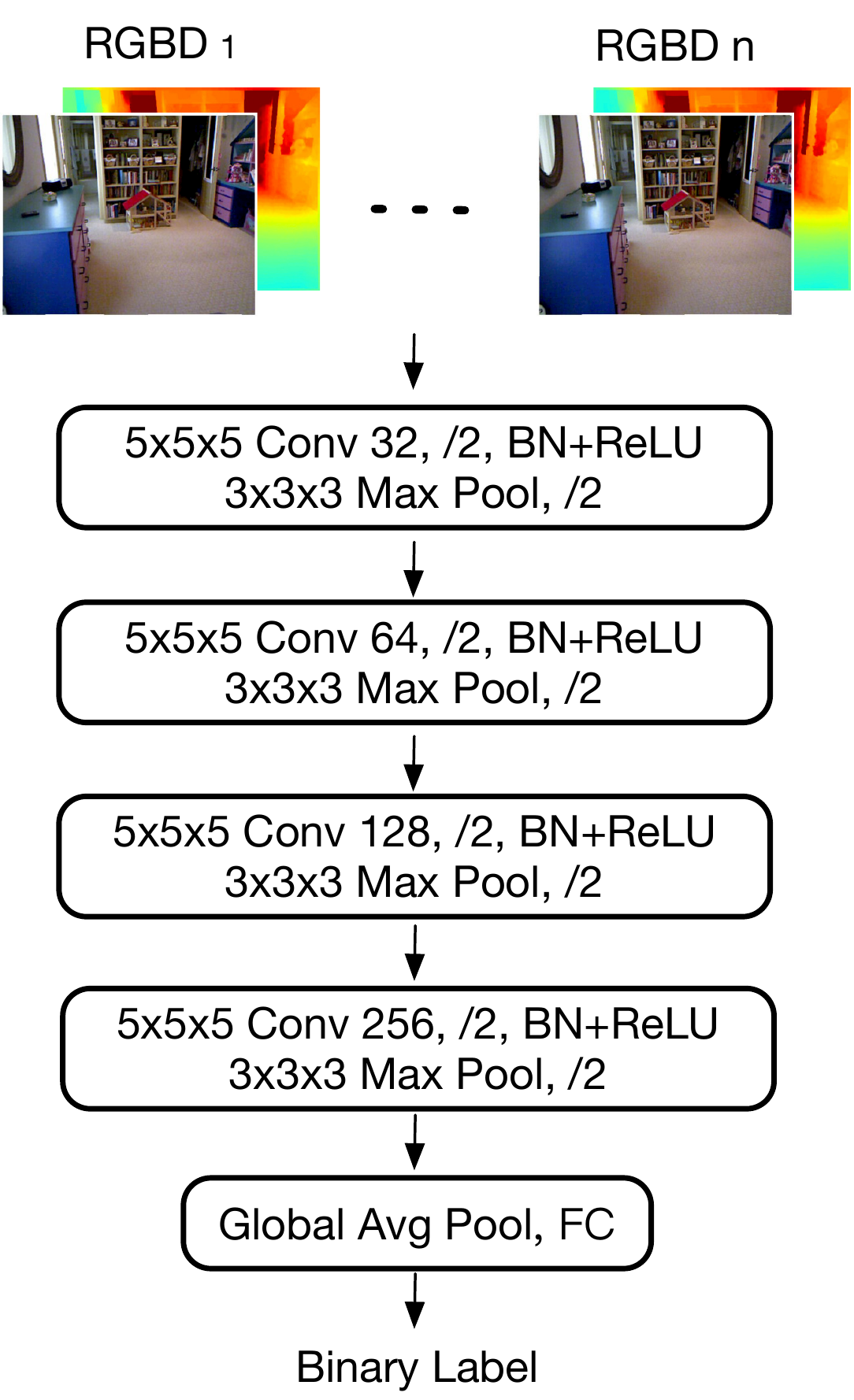}
\end{center}
\vspace{-0.4cm}
   \caption{Structure of the 3DCNN discriminator model in adversarial learning. It contains four convolution blocks, a global average pooling layer and a fully connected layer. It takes as input concatenated RGB-D video frames and output a binary label which indicates the input source.}
\label{fig:3DCNN_struct}
\end{figure}

\subsubsection{Temporal loss}
Our proposed ST-CLSTM is able to exploit the temporal correlations among consecutive video frames. In order to further enforce the consistency among frames, we apply the generative adversarial learning scheme and design a temporal consistency loss. Specifically, after our ST-CLSTM produces depth estimations, we introduce a three-dimensional convolutional neural network (3D CNN) which takes as input the estimated depth sequence and output a score. This score represents the probability of the depth sequence comes from our ST-CLSTM rather than the ground-truths. The 3D CNN is then act as a discriminator. We train the discriminator by maximizing the probability of assigning the correct label to both the estimated and ground-truth depth sequences. Our ST-CLSTM acts as the generator. 
The discriminator tries to distinguish the generator's output (labelled as `fake') from the ground truth depth sequence (labelled as `real'). Upon convergence we wish that the generator's output can appear as close as possible to the ground truth so as to confuse the discriminator.  
During the training of discriminator, we train the generator simultaneously. The objective of our generative adversarial learning is expressed as follows:
\begin{equation}
\begin{aligned}
& \min_{G}\max_{D}V(G,D)=\\
& \mathbb{E}_{\mathbf{z}\in\zeta}[\log(D(\mathbf{z}))]+\mathbb{E}_{\mathbf{x}\in\chi}[\log(1-D(G(\mathbf{x})))],
\end{aligned}
\label{formula:temporal_loss}
\end{equation}
where $\mathbf{x}=[x_1,...x_n]$ are the input RGB frames and $\mathbf{z}=[d_1,...d_n]$ are the ground-truth depth frames. $\chi$ and $\zeta$ are the distributions of input RGB frames and ground-truth depths respectively.

Since our discriminator is a binary classifier, we train it using the cross entropy loss. The cross entropy loss then acts as our temporal loss function. During the training of our ST-CLSTM, we combine our temporal loss with the aforementioned spatial loss as follows:
\begin{equation}
L = {L}_{spatial} + \alpha{L}_{temporal},
\end{equation}
where $\alpha$ is a weighting coefficient. We empirically set it to $0.1$.

The detailed structure of our 3DCNN is illustrated in Fig.~\ref{fig:3DCNN_struct}. It is composed of 4 convolution blocks, a global average pooling layer and a fully-connected layer. Each convolution block contains a 3D convolution layer, followed by a batch normalization layer, a ReLU layer and a max pooling layer. The first 3D convolution layer and all the max pooling layers have a stride of 2. In practice, as plotted in Fig.~\ref{fig:3DCNN_struct}, our 3DCNN takes as input concatenated RGB and depth frames to enforce the consistency between the video frame and the corresponding depth. In order to increase the robustness of our discriminator, in our generated input depth sequences, we randomly mix some ground-truth depth frames with a certain probability. 

Note that, the adversarial training here is mainly to enforce temporal consistency, instead of improving the depth accuracy of single frame's depth as in \cite{Rethinking18GAN}.

\section{Experiments}

In this section, we evaluate our proposed depth estimation framework on the indoor NYU Depth V2 dataset and the outdoor KITTI dataset, and compare against a few existing depth estimation approaches.

\subsection{Datasets}

{\bf NYU Depth V2} contains 464 videos taken from indoor scenes. We apply the same train/test split as in Eigen \emph{et al.}~\cite{eigen2014depth} which contains 249 videos for training, and 654 samples from the rest 215 videos for test. During training, we resize the image from $640\times480$ to $320\times240$ and then crop 
patches of 
$304\times228$ for training.

{\bf KITTI} contains 61 outdoor video scenes captured by cameras and depth sensors mounted on a driving car. We apply the same train/test split as in Eigen \emph{et al.}~\cite{eigen2014depth} which contains 32 videos for training, and 697 samples from the rest 29 videos for test. During training, we randomly crop  patches of size $480\times320$ from the original images as inputs.

\subsection{Evaluation metrics}
{\bf Spatial Metrics} We evaluate the performance of our framework using the commonly applied metrics defined as follows:
1) Mean relative error (Rel): $\frac{1}{N}\sum_{i=1}^{N}{\frac{{||{d}_{i}-{g}_{i}||}_{1}}{{g}_{i}}}$;
2) 
Root mean squared error (RMS): $\sqrt{\frac{1}{N}\sum_{i=1}^{N}{{({d}_{i}-{g}_{i})}^{2}}}$; 
3)
Mean $\log_{10}$ error (log10): $\frac{1}{N}\sum_{i=1}^{N}{{||\log_{10}{d}_{i}-\log_{10}{g}_{i}||}_{1}}$;
4)
Accuracy with threshold t: Percentage of ${d}_{i}$ such that $max(\frac{{d}_{i}}{{g}_{i}}, \frac{{g}_{i}}{{d}_{i}})=\delta < t \in [1.25, {1.25}^{2}, {1.25}^{3}]$. 
$N$ denotes the total number of pixels. $d_i$ and $g_i$ are estimated and ground-truth depths of pixel $i$, respectively.

{\bf Temporal Metrics} Maintaining temporal consistency means keeping the changes and motions among adjacent frames of estimation results consistent with that of corresponding ground truths. In order to quantitatively evaluate the temporal consistency, we introduce two metrics: temporal change consistency (TCC) and temporal motion consistency (TMC). They are defined as:
\def\TCC{ {\rm TCC } }
\def\TMC{ {\rm TMC } }
\def\SSIM{ {\rm SSIM } }
\begin{equation}
\begin{aligned}
& \TCC(D, G)= \\
& \frac{\sum_{i=1}^{n-1}{\SSIM(abs({d}^{i}-{d}^{i+1}), abs({g}^{i}-{g}^{i+1}))}}{n-1},
\end{aligned}
\label{formula:temporal_loss}
\end{equation}

\begin{equation}
\begin{aligned}
& \TMC(D, G)= \\
& \frac{\sum_{i=1}^{n-1}{\SSIM(oflow({d}^{i}, {d}^{i+1}), oflow({g}^{i}, {g}^{i+1}))}}{n-1},
\end{aligned}
\label{formula:temporal_loss}
\end{equation}
where $D=\left({d}^{1}, {d}^{2},\cdots, {d}^{n}\right)$ and $G=\left({g}^{1}, {g}^{2},\cdots, {g}^{n}\right)$ are estimation depth maps of $n$ consecutive frames and the corresponding ground truths. $oflow$ denotes real time $TV-{L}^{1}$ optical flow \cite{zach2007duality}.  SSIM is structural similarity \cite{wang2004image}.

\subsection{Implementation details}
We train our proposed framework for 20 epochs. The initial learning rate of the ST-CLSTM is set to 0.0001 and decrease by a factor of 0.1 after every five epochs. Our spatial feature extraction network in the ST-CLSTM is pretrained on the ImageNet dataset. As for our 3D CNN, the initial learning rate is set to 0.1 for the NYU Depth V2 dataset and 0.01 for the KITTI dataset. The parameters of our 3D CNN are randomly initialized. During the generative adversarial training, before we start to update our 3D CNN parameters, we first train our ST-CLSTM for one epoch for the NYU Depth V2 dataset, and two epochs for the KITTI dataset, to make sure that our ST-CLSTM is able to generate plausible depth estimations.  

Following~\cite{hu2019revisiting}, we employ three data augmentation methods including: 
1) randomly flip the RGB image and depth map horizontally with a probability of 50\%; 
2) rotate the RGB image and depth map by a random degree $c \in[\ang{-5}, \ang{5}]$; 
3) scale the brightness, contrast and saturation values of the RGB image by a random ratio $r \in [0.6, 1.4]$.

\subsection{Benefit of ST-CLSTM}

\newcolumntype{C}[1]{>{\centering\let\newline\\\arraybackslash\hspace{-6pt}}m{#1}}
\begin{table}[h]
\footnotesize
\renewcommand\arraystretch{1.2}
\begin{center}
\begin{tabular}{C{0.04cm}C{1.2cm}|C{0.6cm}C{0.6cm}C{0.6cm}C{0.6cm}C{0.6cm}C{0.6cm}}
\hline
\#  &  model &  Rel & RMS  & log10 & ${\delta}_{1}$ & ${\delta}_{2}$ & ${\delta}_{3}$\\
\hline
\multicolumn{8}{c}{NYU-Depth V2} \\
\hline
  1  & 2DCNN         & 0.139  & 0.585 & 0.059  & 0.819  & 0.961  & 0.990 \\
  3  & ST-CLSTM      & 0.134  & 0.581 & 0.058  & 0.824  & 0.965  & 0.991 \\
  4  & ST-CLSTM      & 0.133  & 0.577 & 0.057  & 0.831  & 0.963  & 0.990 \\
  5  & ST-CLSTM      & {\bf0.132}  & {\bf0.572} & {\bf0.057}  & {\bf0.833}  & {\bf0.966}  & {\bf0.991} \\
\hline
\multicolumn{8}{c}{KITTI} \\
\hline
 1   & 2DCNN         & 0.111  & 4.385 &   0.048  &  0.871     &  0.962    &  0.987   \\
 5   & ST-CLSTM      & {\bf0.104}  & {\bf4.139} & {\bf0.045}   &  {\bf0.883}    & {\bf0.967}    &{\bf0.988}  \\
\hline
\end{tabular}
\end{center}
\vspace{-0.2 cm}
\caption{Experiment results of our ST-CLSTM. The first 4 rows are the results on the NYU Depth V2 dataset and the last 2 rows are the results on the KITTI dataset. \# denotes the number of input frames. ${\delta}_{i}$ means $\delta < {1.25}^{i}$}
\vspace{-0.2 cm}
\label{Table:ST-CLSTM}
\end{table}

The ST-CLSTM is the key component in our proposed depth estimation framework as it captures both spatial and temporal information. In this section, we evaluate the performance of our ST-CLSTM on both indoor and outdoor datasets. The results are reported  in Table~\ref{Table:ST-CLSTM}. We denote the baseline approach that captures no temporal information as 2DCNN. Specifically, we replace the CLSTM in our ST-CLSTM structure with 3 convolution layers. The number of channels are 128, 128 and 1 respectively. Since the temporal information exists among consecutive frames, the number of input frames influences the performance of our ST-CLSTM. We first evaluate the performance of our ST-CLSTM on the NYUD Depth V2 dataset with different number of input frames and show the results in the first 4 rows in Table~\ref{Table:ST-CLSTM}. We can see that with the number of frame increases, the performance increases, as our ST-CLSTM captures more temporal information. 
We use 5 input frames in our experiments  considering the computation cost.

We can see from Table~\ref{Table:ST-CLSTM} that our ST-CLSTM is able to capture the temporal information and improve the depth estimation performance on both indoor and outdoor datasets.

\subsection{Benefit of generative adversarial learning}

\newcolumntype{C}[1]{>{\centering\let\newline\\\arraybackslash\hspace{-3pt}}m{#1}}
\begin{table}[bth]
\footnotesize
\renewcommand\arraystretch{1.2}
\begin{center}
\begin{tabular}{C{1.4cm}|C{0.6cm}C{0.7cm}C{0.8cm}C{0.6cm}C{0.6cm}C{0.6cm}}
\hline
model &  Rel & RMS  & log10 & ${\delta}_{1}$ & ${\delta}_{2}$ & ${\delta}_{3}$ \\
\hline
\multicolumn{7}{c}{NYU-Depth V2} \\
\hline
ST-CLSTM  & 0.132  & 0.572  & 0.057  & 0.833   & {\bf0.966}  & 0.991 \\
GAN      & {\bf0.131}  & {\bf0.571}  & {\bf0.056}  & {\bf0.833}   & 0.965  & {\bf0.991} \\  
\hline
\multicolumn{7}{c}{KITTI} \\
\hline
ST-CLSTM    & 0.104  & 4.139 & 0.045   &  0.883    & 0.967    &0.988  \\
GAN         & {\bf0.101}  & {\bf4.137}  &  {\bf0.043}   &  {\bf0.890}  & {\bf0.970}   &  {\bf0.989}   \\

\hline
\end{tabular}
\end{center}
\vspace{-0.2 cm}
\caption{Experiment results of our generative adversarial learning. The first 2 rows are the results on the NYU Depth V2 dataset and the last 2 rows are the results on the KITTI dataset.}
\vspace{-0.3 cm}
\label{Table:GAN}
\end{table}

In this section, we evaluate the performance of our generative adversarial learning scheme which further enforces the temporal consistency among video frames. The evaluation results on the NYU Depth V2 and the KITTI dataset are %
reported 
in Table~\ref{Table:GAN}. For each dataset, we show the results of our ST-CLSTM 
without and with 
generative adversarial learning, denoted as ST-CLSTM and GAN respectively. We can see from Table~\ref{Table:GAN} that our generative adversarial learning and temporal loss can enforce the temporal consistency and further improve the performance of our ST-CLSTM. 

\subsection{Improvement of temporal consistency}

\begin{figure}[bt]
\begin{center}
\includegraphics[width=3.2in]{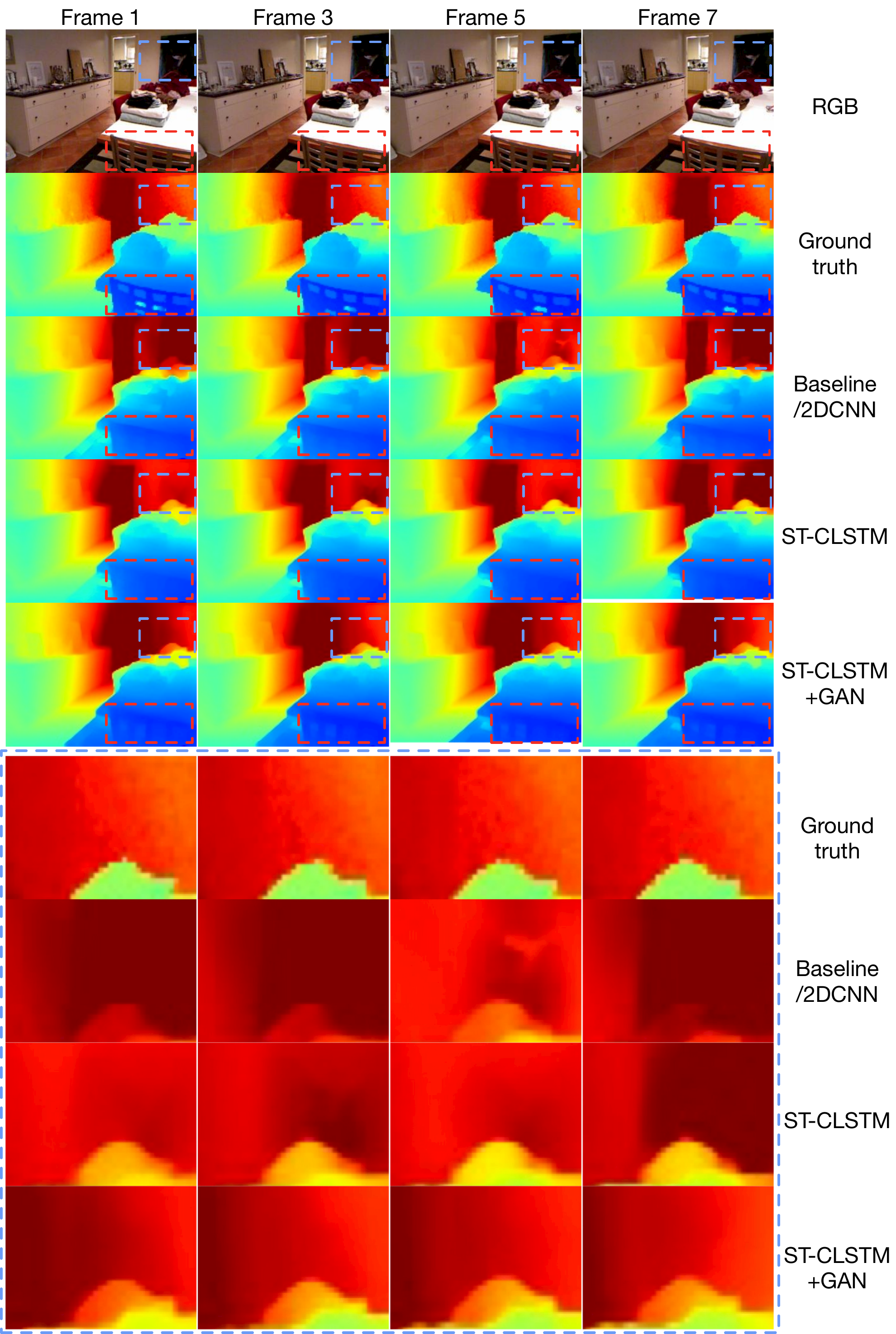}
\end{center}
\vspace{-0.35 cm}
   \caption{Visual results of depth estimation on the NYU Depth V2 dataset. The top five rows are: RGB inputs, ground truth, the results of baseline, ST-CLSTM and ST-CLSTM+GAN. 
   For better visualization, 
   we present the corresponding zoom-in regions of ground truth and estimations results on the four bottom rows. Here, both ST-CLSTM and ST-CLSTM+GAN are trained with 5 frames inputs. From the results on the last row, we can see that the estimation results generated by ST-CLSTM+GAN exhibit  better temporal consistency than that of 2DCNN and ST-CLSTM.}
\vspace{-0.3cm}
 
\label{fig:demo2}
\end{figure}

The major contribution of our work is to exploit temporal information for accurate depth estimation. The aforementioned experiments have revealed that our proposed ST-CLSTM and generative adversarial learning scheme are able to better capture the temporal information and improve the depth estimation performance. In this section, we show the improvement of our proposed framework in the temporal dimension with both visual effects and temporal consistency metrics.

We show the estimated depths of four consecutive frames with one frame gap between each frame in Fig.~\ref{fig:demo2}. We first show the RGB frames and the ground-truth depth maps in the first two rows, then we show the depth estimations of the baseline method (2DCNN) and our proposed framework in the last three rows. 

We highlight a front area and a background area in blue and red dotted windows respectively, and we maximize the blue dotted window for better visualization. Since the four frames are consecutive, the ground-truth depths in these four frames change smoothly. However, the baseline method fails to maintain the smoothness. The estimated depths vary largely. Our ST-CLSTM captures the temporal correlations and produces visually better performance as demonstrated  in Fig.~\ref{fig:demo2}. For all the frames, the edges of objects are sharper and the backgrounds are smoother. With our proposed generative adversarial learning scheme, the temporal consistency is enforced and the performance is further improved. The details are well maintained in all the frames. For instance, the bars of the chair in the red dotted window.\footnote{%
Readers may refer to the demonstration video: \url{https://youtu.be/B705k8nunLU}
}

3D CNN can capture the change and motion information between consecutive frames, as it convolves the input along both the spatial and temporal dimensions. To confuse  the 3D CNN discriminator, the change and motion of estimation results must keep consistent with that of corresponding ground truths.
We sampled 654 sequences from test set with a length of 16 frames each and report the average TCC and TMC in Table~\ref{Table:TCC_TMC}, from which we can see that the 3D CNN discriminator does not only improve the estimation accuracy, but also better enforces the temporal consistency.
\newcolumntype{C}[1]{>{\centering\let\newline\\\arraybackslash\hspace{-12pt}}m{#1}}
\begin{table}[t]
\footnotesize
\renewcommand\arraystretch{1.2}
\begin{center}
\begin{tabular}{C{0.95cm}|C{0.75cm}C{0.4cm}C{0.4cm}C{0.4cm}C{0.4cm}C{0.4cm}C{0.5cm}C{0.5cm}}
\hline
Model &  Rel & RMS  & log10 & ${\delta}_{1}$ & ${\delta}_{2}$ & ${\delta}_{3}$ & TCC & TMC \\
\hline
Baseline  & 0.139  & 0.585  & 0.059  & 0.819  & 0.961  & 0.990  & {}0.846  &  0.956 \\
ST-CLSTM  & 0.132  & 0.572  & 0.057  & 0.833  & 0.966  & 0.991  & 0.866  &  0.962 \\
3D-GAN    & 0.131  & 0.571  & 0.056  & 0.833  & 0.965  & 0.991  & {\bf0.870}  &  {\bf0.965} \\
\hline
\end{tabular}
\end{center}
\caption{Experiment results on NYU Depth V2.}
\vspace{-0.5cm}
\label{Table:TCC_TMC}
\end{table}

\newcolumntype{C}[1]{>{\centering\let\newline\\\arraybackslash\hspace{-3pt}}m{#1}}
\begin{table*}
\footnotesize
\renewcommand\arraystretch{1.2}
\begin{center}
{
\begin{tabular}{rccccccccccccccccccccccccc}
\hline
Method  & Rel & RMS$\downarrow$  & log10 &  ${\delta}_{1}$ & ${\delta}_{2}$ & ${\delta}_{3}$ & backbone \\
\hline\hline
DepthTransfer~\cite{karsch2014depth}         & 0.350    & 1.200    & 0.131    & -        & -        & -     & - \\ 
Make3D~\cite{saxena2009make3d}          & 0.349    & 1.214    & -        & 0.447    & 0.745    & 0.897 & - \\
Liu \emph{et al.}~\cite{liu2014discrete}      & 0.335    & 1.060    & 0.127    & -        & -        & -     & - \\
Li \emph{et al.}~\cite{li2015depth}       & 0.232    & 0.821    & 0.094    & 0.621    & 0.886    & 0.968 & - \\
Liu \emph{et al.}~\cite{liu2015deep}      & 0.230    & 0.824    & 0.095    & 0.614    & 0.883    & 0.971 & - \\
Wang \emph{et al.}~\cite{wang2015towards}     & 0.220    & 0.824    & -        & 0.605    & 0.890    & 0.970 & - \\
Liu \emph{et al.}~\cite{liu2016learning}      & 0.213    & 0.759    & 0.087    & 0.650    & 0.906    & 0.976 & - \\
Eigen \emph{et al.}~\cite{eigen2015predicting}    & 0.158    & 0.641    & -        & 0.769    & 0.950    & 0.988 & - \\
Chakrabarti \emph{et al.}~\cite{chakrabarti2016depth}    & 0.149    & 0.620    & -        & 0.806    & 0.958    & 0.987 & VGG\-19 \\
Li \emph{et al.}~\cite{li2017two}       & 0.143    & 0.635    & 0.063    & 0.788    & 0.958    & 0.991 & VGG\-16 \\
Ma \& Karaman~\cite{mal2018sparse}       & 0.143    &  -       & -        & 0.810    & 0.959    & 0.989 & ResNet-50 \\
Laina \emph{et al.}~\cite{laina2016deeper}    & 0.127    & 0.573    & 0.055    & 0.811    & 0.953    & 0.988 & ResNet\-50 \\
Pad-net~\cite{xu2018pad}       & 0.120    & 0.582    & 0.055    & 0.817    & 0.954    & 0.987 & ResNet\-50 \\
DORN ~\cite{fu2018deep}          &{\bf0.115}&{\bf0.509}&{\bf0.051}& 0.828    & 0.965 &{\bf0.992}  & ResNet\-101 \\
\hline
Ours            & 0.131    & 0.571    & 0.056    &{\bf0.833}&{\bf0.965}& 0.991   & ResNet\-18 \\

\hline
\end{tabular}
}
\end{center}
\vspace{-0.2cm}
\caption{Comparisons with state-of-the-arts on the NYU Depth V2 dataset. We show our results in the last row.}
\vspace{-0.35 cm}
\label{Table: NYUv2}
\end{table*}

\newcolumntype{C}[1]{>{\centering\let\newline\\\arraybackslash\hspace{-3pt}}m{#1}}
\begin{table*}
\footnotesize
\renewcommand\arraystretch{1.2}
\begin{center}
\begin{tabular}{rcccccccccccccccccccccccc}
\hline 
Method & Rel & RMS  & log10  & ${\delta}_{1}$ & ${\delta}_{2}$ & ${\delta}_{3}$ &  backbone \\
\hline\hline
Make3D~\cite{saxena2009make3d}                           & 0.280    & 8.734    &    -  & 0.601    & 0.820    & 0.926 & -\\
Eigen \emph{et al.}~\cite{eigen2014depth}                & 0.190    & 7.156    &    -  & 0.692    & 0.899    & 0.967 & -\\
Liu \emph{et al.}~\cite{liu2016learning}                  & 0.217    & 6.986    &    -  & 0.647    & 0.882    & 0.961 & - \\
LRC~\cite{godard2017unsupervised}                        & 0.114    & 4.935    &    -  & 0.861    & 0.949    & 0.976  & ResNet-50 \\
Kuznietsov \emph{et al.} \cite{kuznietsov2017semi}        & 0.113    & 4.621    &    -  & 0.862    & 0.960    & 0.986  & ResNet-50 \\
\hline
Mahjourian \emph{et al.} \cite{mahjourian2018unsupervised}& 0.159    & 5.912    &    -  & 0.784    & 0.923    &  0.970 & DispNet~\cite{mayer2016large} \\
Zhou \emph{et al.}~\cite{zhou2018unsupervised}           & 0.143    & 5.370    &    -  & 0.824    & 0.937    &  0.974 & VGG-19 \\
\hline
Ours                    & {\bf0.101}    & {\bf4.137}    & {\bf0.043}  & {\bf0.890}    & {\bf0.970}    & {\bf0.989}  & ResNet-18 \\

\hline
\end{tabular}
\end{center}
\vspace{-0.25cm}
\caption{Comparisons with state-of-the-art methods on the KITTI dataset. The first five rows are the results of static image depth estimation methods.
The following two rows are the results of video depth estimation methods, and the last row are our results.}
\vspace{-0.4 cm}
\label{Table: KITTI}
\end{table*}

\subsection{Comparisons with state-of-the-art results}
In this section, we evaluate our approach on the NYU Depth V2 dataset and the KITTI dataset and compare with some state-of-the-art results. The results are reported  in Table~\ref{Table: NYUv2} and Table~\ref{Table: KITTI} respectively. We can see that with our captured temporal information, we outperform most  state-of-the-art methods which often use  more complicated network structures. The aim of our work is to exploit temporal information for real-time depth estimation. We apply a shallow ResNet18 model as our backbone. The performance of our approach can be improved with deeper backbone networks. We leave this as future work.

\subsection{Speed analysis}

\newcolumntype{C}[1]{>{\centering\let\newline\\\arraybackslash\hspace{-3pt}}m{#1}}
\begin{table}[h]
\footnotesize
\renewcommand\arraystretch{1}
\begin{center}
\begin{tabular}{rccc}
\hline
 Model  &  Dataload   & Time (ms per frame)  &  Speed (fps) \\
\hline
 Baseline    & S-mode    &        28.90           &   34.60      \\
 ST-CLSTM    & S-mode    &        30.22           &   33.09      \\
 ST-CLSTM    & PS-mode   &        5.72            &   174.83     \\

\hline
\end{tabular}
\end{center}
\vspace{-0.2 cm}
\caption{Processing speed of different models and data loading modes. The resolution of input frame is $304\times 228$.
}
\vspace{-0.8 cm}
\label{Table: speed}
\end{table}

One of the contributions of our work here is that our model can execute in real-time for practical applications. In this section, we evaluate the processing time of our model. Specifically, we feed our model videos with spatial resolution of $304\times228$. We test 600 frames for five epochs and report the mean values. We load the videos in two different ways: 1) Serial mode (S-mode). We load the video frames one by one. 2) Parallel+serial mode (PS-mode). We feed 120 frames to our spatial extraction network at one time to obtain the spatial features, then we feed the spatial features to our CLSTM one by one.

We implement our model with the PyTorch~\cite{paszke2017automatic}, and perform the inference on a computer with 8GB RAM,  Intel i7-4790 CPU and  GTX1080Ti GPU. We report the processing time of one frame, and the frame rate in Table~\ref{Table: speed}. We can see that compared with the baseline (2D CNN) method, our ST-CLSTM   method shows negligible drop of processing speed. Moreover, when we adopt the PS-mode for data loading, our processing speed increases dramatically. As the frame rate of common video formats is less than 30fps, our model is sufficiently fast  to work in real-time.

\section{Conclusion}

In this work,
we have proposed a novel ST-CLSTM structure by combining a shallow 2D CNN and a CLSTM. Our ST-CLSTM is able to capture both spatial features and temporal correlations among video frames for depth estimation. We have also designed a novel temporal loss by introducing the generative adversarial learning scheme. Our temporal loss is able to further enforce temporal consistencies among video frames. Experiments on benchmark indoor and outdoor datasets reveal that our proposed framework can effectively capture temporal information and achieve outstanding performance. Moreover, our proposed framework is able to execute in real-time for real-world applications, and can be easily generalized to most existing depth estimation frameworks.

{\bf Acknowledgments} 
We would like to thank Huawei Technologies for the donation of  GPU cloud computing resources. This work was  in part supported by the National Natural Science Foundation of China (61871460, 61876152), Fundamental Research Funds for the Central Universities (3102019ghxm016) and Innovation Foundation for Doctor Dissertation of Northwestern Polytechnical University (CX201816).

{\small
\bibliographystyle{ieeetr}
\bibliography{main}
}

\end{document}